# On the Sample Complexity of Learning Bayesian Networks

**Nir Friedman**
Stanford University
Dept. of Computer Science
Gates Building 1A
Stanford, CA 94305-9010
nir@cs.stanford.edu

**Zohar Yakhini**
Stanford University
Dept. of Mathematics
Stanford, CA 94305
zohary@gauss.stanford.edu

## Abstract

In recent years there has been an increasing interest in learning *Bayesian networks* from data. One of the most effective methods for learning such networks is based on the *minimum description length* (MDL) principle. Previous work has shown that this learning procedure is asymptotically successful: with probability one, it will converge to the target distribution, given a sufficient number of samples. However, the rate of this convergence has been hitherto unknown.

In this work we examine the *sample complexity* of MDL based learning procedures for Bayesian networks. We show that the number of samples needed to learn an $\epsilon$-close approximation (in terms of entropy distance) with confidence $\delta$ is $O\left(\left(\frac{1}{\epsilon}\right)^{\frac{4}{3}} \log \frac{1}{\epsilon} \log \frac{1}{\delta} \log \log \frac{1}{\delta}\right)$. This means that the sample complexity is a low-order polynomial in the error threshold and sub-linear in the confidence bound. We also discuss how the constants in this term depend on the complexity of the target distribution. Finally, we address questions of asymptotic minimality and propose a method for using the sample complexity results to speed up the learning process.

## 1 Introduction

*Bayesian networks* (BN) are the representation of choice for uncertainty in artificial intelligence. This representation allows to compactly represent a joint distribution of several random variables. The key idea of BNs is the explicit representation of (in)dependencies in the distribution. These independencies are exploited to compactly represent numerical parameters and for efficient inference.

In recent years there has been a growing number of results on learning BNs from data. This problem is crucial in many applications in a wide range of domains. Recent results focused on the theoretical development of learning procedures and on empirical methods of learning. However, the complexity of the learning problem is generally unknown (an exception is [Höffgen 1993], which we discuss below).

A BN is composed of two parts. The first part is a directed acyclic graph that represents (in)dependencies among the random variables: $X$ is independent of $Y$ given $Z$, if $Z$ *separates* (under the appropriate definition) $X$ from $Y$ in the graph. The second part consists of numerical parameters. These parameters encode the conditional probability of each variable given its parents in the network.

The learning task can be specified as follows. Given a database of independently drawn instances (each of which assigns values to all the variables in our domain), construct a BN that best describes the joint distribution in the database. This problem is usually decomposed into two tasks. The first task is learning the best parameters for a fixed network structure. This problem is usually solved by choosing the parameters that maximize the likelihood of the data.[1] This maximization has a simple closed form solution (see Section 2). The second task involves learning the best network structure. This is a discrete search problem. We have to define a measure of quality for each network candidate and search for a structure with maximal quality. It turns out that the likelihood measure is not suitable for learning structures. It is straightforward to see that more complicated network structures (i.e., ones that make less assumptions of independence) are favored by this measure. Such complex structures, however, are undesirable since they fail to capture (and exploit) independencies in the domain and tend to overfit the training data.

One possible solution to this problem is to use the *minimum description length* (MDL) metric [Lam and Bacchus 1994; Suzuki 1993], which adds a *penalty term* to the likelihood measure. This term penalizes complex networks, i.e., networks that embody large numbers of parameters. The size of the penalty depends on the exact number of parameters and a *penalty weight* that might depend on the sample size. We examine a range of penalty weight functions: constants (do not depend on the sample size), logarithmic (the standard choice in the literature), and polynomial.

The introduction of this penalty term makes it more likely that the highest scoring network is a simple one. This raises the question of whether these MDL scores learn the true target distribution. Barron and Cover [1991] show that, in a very general setting, MDL methods are asymptotically successful: assuming that the training samples are drawn independently from a fixed *target distribution*, then with probability 1, given a large enough number of samples, networks that describe the target distribution will score strictly better than all others. However, the rate of convergence (in probability) has been hitherto unknown.

---

[1] A slightly more complicated variant occurs in *Bayesian* learning, where, roughly speaking, we choose the parameters with highest posterior value [Heckerman, Geiger, and Chickering 1995].

In this paper we examine the *sample complexity* of learning Bayesian networks. Roughly speaking, the sample complexity measures how many instances are needed to learn a Bayesian network from data. The setup is as follows. We assume that there is a Bayesian network $B^*$ that describes the target distribution from which training samples are drawn. Clearly it is not always necessary nor possible to learn the exact target distribution. Instead a close approximation suffices. In this paper we use the *entropy distance* (also known as the *Kullback-Leibler distance*) [Kullback and Leibler 1951] as the measure of approximation:

$$D(P\|Q) = \sum_x P(x) \log \frac{P(x)}{Q(x)}.$$

This measure of distance is accepted as a standard measure of error in the Bayesian networks literature [Heckerman, Geiger, and Chickering 1995; Lam and Bacchus 1994; Pearl 1988]. (See Section 3.1 for a detailed motivation of this choice.) It is also clear that since we are learning from random samples, there is some probability of seeing unrepresentative sequences that might mislead us. This probability, however, decreases with $N$.

To specify the acceptable learning criteria we need to set two parameters. The first is the *error threshold*, which we will denote by $\epsilon$, and the second is the *confidence threshold*, which we denote by $\delta$. We are interested in constructing a function $N(\epsilon, \delta)$, such that if the sample size is larger than $N(\epsilon, \delta)$, then

$$\Pr(D(P_{B^*}\|P_{Lrn()}) > \epsilon) < \delta.$$

where $Lrn()$ represents the learning routine, i.e., it is a function of the training data that returns a network. (Note that the probability here is the one corresponding to independently sampling $P_{B^*}$, see Section 3.) If the function $N(\epsilon, \delta)$ is minimal, in the sense that it always return the smallest $N$ that satisfies this requirement, then it is called the *sample complexity*. We examine upper-bound results on the behavior of the sample complexity of the MDL-based learning routine.

As we will show, the sample complexity depends on two different factors. The first is the interaction between the likelihood measure and the penalty term. The second is the noise introduced by the random nature of the sampling process. We start by examining the first interaction: We assume that the sample is *ideal*, i.e., it exactly reflects $P_{B^*}$. This allows us to give a simple description of the interaction between the likelihood and the penalty as the sample size grows. In the second stage we examine how the random sampling process affects the learning procedure. This random process introduces errors in the estimation of the likelihood of candidate structures. We show how to bound the probability that this error will cause us to prefer a "bad" candidate over a good one.

Our main result can be summarized as follows: there is a function

$$N(\epsilon, \delta) \in O\left((\frac{1}{\epsilon})^{\frac{4}{3}} \log \frac{1}{\epsilon} \log \frac{1}{\delta} \log \log \frac{1}{\delta}\right)$$

such that if $N > N(\epsilon, \delta)$, then the probability of learning, using the standard logarithmic penalty weight, a network that is $\epsilon$-close to the target distribution, is greater than $1 - \delta$. This means that the sample complexity is a low-order polynomial in $\frac{1}{\epsilon}$ and sub-linear in $\frac{1}{\delta}$. This is encouraging, since it shows that the number of samples necessary to improve the accuracy of the approximation grows at a reasonable rate.

Our analysis also takes into account the complexity of the target distribution, which affects the constants involved in the asymptotic statement above. More precisely, let $B^*$ denote the (minimal) BN which describes the underlying distribution. One measure of the complexity is the number of parameters in $B^*$. We show that the sample complexity is sub-linear in the number of parameters of $B^*$. The second measure of complexity is the *skewness* of $B^*$, i.e., how close to the boundary of the probability simplex the underlying distribution is. Intuitively, a skewed target distribution is harder to learn, since we need many samples to assess the probability of unlikely events. We take the measure of skewness to be the smallest non-zero probability of a primitive event. We show that the sample complexity is a low-order polynomial in this quantity.

Our analysis also examines the effect of choosing different penalty weight functions. As expected, we observe that large penalty weights lead to slower convergence. However, our upper-bound result does not distinguish between weights that grow slower than the logarithm of the number of samples. This suggests that a logarithmic penalty weight does not slow down the convergence rate of the learning procedure.

We also examine behavior in the limit. As mentioned above it is well known that MDL scoring metric is *consistent*, i.e., converges almost surely to the right distribution. This does not address the question of *asymptotic minimality*, i.e., whether the learned network structure is one of the minimal BNs that describe the target distribution. We show that for some penalty weights the process is indeed asymptotically minimal.

Finally, we examine a slightly different topic. We describe how, using sample complexity results, we can speed up the learning process. The rough idea is to use tight estimates of the likelihood of candidate networks, without examining all the training data. Instead, we suggest a sub-sampling strategy that examines portions of the data which are large enough to evaluate the candidate.

The paper is organized as follows. In Section 2 we briefly overview Bayesian networks and MDL procedures for learning them. In Section 3 we examine the sample complexity of learning using MDL. In Section 4 we briefly discuss sub-sampling strategies for learning BNs. Finally, in Section 5 we discuss related work, possible improvements, and future directions of research. Detailed proofs are in the extended version of this paper [Friedman and Yakhini 1996].

## 2  Learning Bayesian Networks

Consider a vector valued random variable, $\mathbf{U} = \{X_1, \ldots, X_n\}$, where each component $X_i$ may take on values from a finite domain. We use capital letters, such as $X, Y, Z$, for variable names and lowercase letters $x, y, z$ to denote specific values taken by those variables. Vector valued variables are denoted by boldface capital letters $\mathbf{X}, \mathbf{Y}, \mathbf{Z}$, and assignments of values to these, by boldface lowercase letters $\mathbf{x}, \mathbf{y}, \mathbf{z}$. The set of values $\mathbf{U}$ can attain is denoted $val(\mathbf{U})$, and the cardinality of $val(\mathbf{U})$ is denoted $||\mathbf{U}||$. Finally, let $P$ be a joint probability distribution over the variables in $\mathbf{U}$, and let $\mathbf{X}, \mathbf{Y}, \mathbf{Z}$ be subsets of $\mathbf{U}$. $\mathbf{X}$ and $\mathbf{Y}$ are *conditionally independent* given $\mathbf{Z}$ if for all $\mathbf{x} \in val(\mathbf{X}), \mathbf{y} \in val(\mathbf{Y}), \mathbf{z} \in val(\mathbf{Z})$, $P(\mathbf{x} \mid \mathbf{z}, \mathbf{y}) = P(\mathbf{x} \mid \mathbf{z})$ whenever $P(\mathbf{y}, \mathbf{z}) > 0$.

A *Bayesian network* is an annotated directed acyclic graph that encodes a joint probability distribution of a domain composed of a set of random variables. Formally, a Bayesian network for $\mathbf{U}$ is the pair $B = \langle G, \Theta \rangle$. $G$ is a directed acyclic graph whose nodes correspond to the random variables $X_1, \ldots, X_n$, and whose edges represent direct dependencies between the variables. The graph structure $G$ encodes the following set of independence assumptions: each node $X_i$ is independent of its non-descendants given its parents in $G$ [Pearl 1988]. The second component of the pair, namely $\Theta$, represents the set of parameters that quantifies the network. It contains a parameter $\theta_{x_i|\Pi_{x_i}} = P(x_i|\Pi_{x_i})$ for each possible value $x_i$ of $X_i$ and $\Pi_{x_i}$ of $\Pi_{X_i}$ (the set of parents of $X_i$ in $G$). $B$ defines a unique joint probability distribution over $\mathbf{U}$ given by

$$P_B(X_1, \ldots, X_n) = \prod_{i=1}^n P_B(X_i|\Pi_{X_i}) = \prod_{i=1}^n \theta_{X_i|\Pi_{X_i}}. \quad (1)$$

Finally we note that a Bayesian network is assumed to be *minimal* in the sense that removing any edge from $G$ would introduce independencies that do not hold in $P_B$ [Pearl 1988].

The problem of learning a Bayesian network can be stated as follows. Given a *training set* $\omega_N = \{\mathbf{u}_1, \ldots, \mathbf{u}_N\}$ of instances of $\mathbf{U}$, find a network $B$ that *best matches* $D$. To formalize the quality of a network's matching to the data, we normally introduce a scoring function. To solve the optimization problem we usually rely on heuristic search techniques over all possible structure candidates.

Let $B = \langle G, \Theta \rangle$ be a Bayesian network, and let $\omega_N = \langle \mathbf{u}_1, \ldots, \mathbf{u}_N \rangle$ (where each $\mathbf{u}_i = \langle x_1^{(i)}, \ldots, x_n^{(i)} \rangle$ assigns values to all the variables in $\mathbf{U}$) be the training sequence of instances. We assume that the $\mathbf{u}_i$ are sampled independently of each other, from the same distribution. The *log-likelihood* of $B$ given $\omega_N$ is simply the log of the probability, according to $B$, of sampling $\omega_N$:

$$LL_{\omega_N}(B) = \log P_B(\mathbf{u}_1, \ldots, \mathbf{u}_N) = \sum_{j=1}^N \log(P_B(\mathbf{u}_j)).$$

The higher this value is, the closer $B$ is to modeling the probability distribution in the data $D$. Using (1) we can decompose the log-likelihood according to the structure of the network. We start with some definitions. Let $\omega_N$ be a training sequence. We define the *empirical distribution* $\hat{P}_{\omega_N}$ to be the frequency function in $\omega_N$:

$$\hat{P}_{\omega_N}(A) = \frac{1}{N} \sum_j 1_A(\mathbf{u}_j) \text{ where } 1_A(\mathbf{u}) = \begin{cases} 1 & \text{if } \mathbf{u} \in A \\ 0 & \text{if } \mathbf{u} \notin A. \end{cases}$$

Then, we can easily derive the following equation:

$$\begin{aligned} LL_{\omega_N}(B) &= N \sum_{\mathbf{u} \in val(\mathbf{U})} \hat{P}_{\omega_N}(\mathbf{u}) \log P_B(\mathbf{u}) \quad (2) \\ &= N \sum_{\mathbf{u} \in val(\mathbf{U})} \hat{P}_{\omega_N}(\mathbf{u}) \sum_i \log \theta_{x_i|\Pi_{x_i}} \\ &= N \sum_i \sum_{x_i, \Pi_{x_i}} \hat{P}_{\omega_N}(x_i, \Pi_{x_i}) \log \theta_{x_i|\Pi_{x_i}} \quad (3) \end{aligned}$$

Now assume that we fix the structure of the network $G$, and want to choose $\Theta$ to maximize $LL_{\omega_N}(\langle G, \Theta \rangle)$. Standard arguments show how to choose the best parameters: We define $\Theta_{G,\omega_N}$ to be the parameter assignment to $G$ given by

$$\theta_{x_i|\Pi_{x_i}} = \hat{P}_{\omega_N}(x_i|\Pi_{x_i}). \quad (4)$$

We denote by $P_{G,\omega_N}$ the probability distribution described by $\langle G, \Theta_{G,\omega_N} \rangle$.

**Lemma 2.1:** *Let* $B = \langle G, \Theta \rangle$ *be a BN. Then* $LL_{\omega_N}(\langle G, \Theta_{G,\omega_N} \rangle) \geq LL_{\omega_N}(B)$.

Thus, in order to maximize the log-likelihood, given a fixed structure, we take the parameters to be the corresponding long-run frequencies in the data.

Since we want to optimize the choice of parameters, we define the log-likelihood of a structure $G$ as the log-likelihood of the best network with this structure:

$$LL_{\omega_N}(G) = LL_{\omega_N}(\langle G, \Theta_{G,\omega_N} \rangle).$$

Using (3) and (4) we observe that:

$$LL_{\omega_N}(G) = -N \sum_i H_{\hat{P}_{\omega_N}}(X_i|\Pi_{X_i}) \quad (5)$$

where $H_P(X|Y) = -\sum_{x,y} P(x,y) \log P(x|y)$ is the *conditional entropy* of $X$ given $Y$, under the distribution $P$.

The log-likelihood score, while very simple, is not suitable for learning the structure of the network, since it favors complete graph structures (in which every variable is connected to every other variable). To see this, suppose that $G \subset G'$ (i.e., every edge in $G$ appears in $G'$). Thus, $\Pi_X^G \subseteq \Pi_X^{G'}$, for every $X$. Using the *data processing inequality* [Cover and Thomas 1991], we get that $H(X|\Pi_X^G) \geq H(X|\Pi_X^{G'})$. It immediately follows that $LL_{\omega_N}(G') \geq LL_{\omega_N}(G)$.

This preference is highly undesirable, since complete structures usually overfit the training data. More precisely, a learning procedure based on the log-likelihood score might learn a BN with many parameters. Unfortunately, the assessment of these parameters from the small number of samples is usually unreliable, and thus leads to poor performance on test data.

One possible way of addressing this problem is to introduce a penalty term for complex structures. We start with a definition. Let $G$ be a network structure. We define $|G|$ to be the number of parameters necessary to quantify a BN with structure $G$. This number corresponds to the statistical *dimension* of a model $\langle G, \Theta \rangle$. Since we argued above that larger networks are less desirable we might consider the following penalized scoring metric:

$$S_\psi(\omega_N, G) = LL_{\omega_N}(G) - |G| \cdot \psi(N)$$

where $\psi(N)$ is a *penalty weight function*. Maximizing the the penalized score involves tradeoffs between the goodness of fit and the number of parameters. Thus, the score of a larger network might be worse than that of a smaller network, even though the former might match the data better.

One useful motivation for this form of a scoring metric is the *minimum description length* (MDL) principle [Rissanen 1989]. The basic idea of the MDL principle is quite simple and appealing. Suppose agent 1 intends to send a file to agent 2 on some communication line. Since the agents want to minimize the number of bits sent, agent 1 should compress the file before transmission. However, since files might come from different sources, a predetermined encoding will be inefficient. Thus, the agents adopt a *two-stage encoding*. In the first stage agent 1 transmits a *model* of the data that implicitly describes an encoding scheme. Then agent 1 transmits the actual data, encoded according to the model. Of course the agents have to agree, in advance, on the class of possible models and on how they are encoded. In order to minimize transmission, agent 1 searches for a model that requires communication: one for which the resulting two-stage encoding is of minimal length. Minimizing this total length involves a tradeoff between the goodness of model and its complexity. In particular, longer sequences of data might justify more complex models.

We do not repeat the exact derivation of the MDL scoring metric for BNs here. The interested reader should consult [Lam and Bacchus 1994]. The general idea is that the class of models agent 1 examines consists of BNs. Each BN describes a probability measure on instances, which defines a Hoffman code, in a unique manner. The description of $\omega_N$ using such a Huffman code requires approximately $-LL_{\omega_N}(G)$ bits. The penalty term simply counts how many bits we need to encode the parameters in a network with structure $G$, where we store $\psi(N)$ bits for each parameter in $\Theta$.[2] Thus, $-S_\psi(\omega_N, G)$ is the total description length of the data when agent 1 chooses to use $G$ as a model for coding $\omega_N$.

Recall that our scoring metric involves a penalty weight function. We consider three classes of penalty weight functions: constants, logarithmic functions, and polynomial functions.

When we set $\psi(N) = c$, we assign a fixed penalty that does not depend on the training set size. This amounts to stating that the complexity of the model is a secondary issue. To see this, note that (5) indicates that the log-likelihood term grows linearly in $N$ and will quickly dominate the penalty term. Thus, the penalty will be taken into account only when comparing two candidates that perform equally well (e.g., have approximately the same log-likelihood). This scoring metric is known as the *Akaike Information Criterion* (AIC) [Akaike 1974]. We note that a constant penalty weight does not constitute a proper MDL encoding scheme, since it assigns a fixed number of bits to the description of the network parameters.

Another possible choice is a logarithmic penalty function. The standard penalty examined in the MDL literature is $\psi(N) = \frac{1}{2} \log N$. The resulting scoring metric is also known as the *Bayesian Information Criterion* (BIC) [Schwarz 1978]. We now discuss in some detail the motivation for this penalty weight.

We start with the MDL approach. Notice that frequencies in $\omega_N$ are fractions with precision of $\frac{1}{N}$. Thus, we need only $\log N$ bits to precisely encode the parameters. However, the central limit theorem says that these frequencies are roughly normally distributed with a variance of $N^{-\frac{1}{2}}$. Hence, the higher $\frac{1}{2} \log N$ bits in the description of the observed fraction are not very useful, and suffices to encode $N^{\frac{1}{2}}$ possible frequency values. Thus, we only need $\frac{1}{2} \log N$ bits to encode each parameter. (We note that this "rounding" operation increases the description length of the data only by a negligible fraction of a bit for each sample.)

The same penalty function also has a justification based on asymptotic approximation of Bayesian statistics. In this alternative approach to learning Bayesian networks, one starts with a *prior* distribution over all possible Bayesian networks. Given the training data $\omega_N$, one can evaluate the *posterior* probability of each network structure $\Pr(G|\omega_N)$. This posterior, of course, depends on the particular prior distribution one chooses. However, for large $N$s, the prior preferences are negligible compared to the evidence in data, as long as the prior does not rule out (i.e., gives probability 0 to) any candidate. It turns out that $S_\psi(\omega_N, G)$, with $\psi(N) = \frac{1}{2} \log N$ is an $O(1)$ approximation of the logarithm of the posterior distribution $\log \Pr(G|\omega_N)$ for well-behaved priors [Schwarz 1978]. We refer the interested reader to [Heckerman, Geiger, and Chickering 1995; Heckerman 1995] for a description of the Bayesian approach to learning Bayesian networks and of Schwarz's result in the context of Bayesian networks.

Finally, we might assign a polynomial penalty, i.e., take $\psi(N) = N^\alpha$ for some $\alpha > 0$. This penalty term is asymptotically larger than any logarithmic penalty. Thus, it embodies a strong bias against large networks.

We now turn to the learning procedure. We assume that the learning procedure searches over the space of all possible graph structures and returns one of the graph structures which scored best. Formally, we define $Lrn_\psi(\omega_N)$ to be a function that returns a network structure $B = \langle G, \Theta_{G,\omega_N} \rangle$, such that

$$S_\psi(\omega_N, Gph(Lrn_\psi(\omega_N))) = \max_G S_\psi(\omega_N, G),$$

where we define $Gph(\langle G, \Theta \rangle) = G$. We note that this learning procedure is unrealistic from a computational standpoint, since the space of network structures is huge, and finding the minimal network structure is an intractable problem [Heckerman 1995]. However, this (standard) idealization allows us to carry out our analysis.

---

[2] The MDL score also has a term for describing the structure of the graph. This term, however, is bounded by $n^2$ and thus can be ignored for large enough $N$.

## 3  Sample Complexity

We now examine the sample complexity of the learning procedure we have just described. From now on, we assume that $B^* = \langle G^*, \Theta^* \rangle$ is a Bayesian network that describes the *target distribution*. We assume that $G^*$ is minimal in the sense that no Bayesian network with a structure $G$ such $|G| < |G^*|$ can describe the target distribution. We assume that the training data $\omega_N$ is a string of $N$ independent samples drawn according to this distribution. We use Pr to denote the probability measure on such sequences (i.e., the product measure $P_{B^*}^{\mathbf{Z}}$).

Recall that we are interested in constructing a function $N(\epsilon, \delta)$ such that if $N > N(\epsilon, \delta)$, then

$$\Pr(\{\omega_N : D(P_{B^*} \| P_{Lrn_\psi(\omega_N)}) > \epsilon\}) < \delta.$$

As we will show, the sample complexity depends on two different factors. The first is the interaction between the log-likelihood scores and the penalty term as $N$ grows larger. The second factor is the random noise introduced by the sampling process. We begin our analysis by isolating the first factor. This is done by assuming that the empirical distribution is *ideal*, i.e., exactly the same as the target distribution. This treatment illustrates the effect of the penalty term on the sample complexity, and prepares the ground for the treatment of the general case.

### 3.1  Entropy Distance

We start by reviewing the choice of entropy distance for measuring error. The *entropy distance* [Kullback and Leibler 1951] is defined as

$$D(P\|Q) = \sum_{\mathbf{x}} P(\mathbf{x}) \log \frac{P(\mathbf{x})}{Q(\mathbf{x})}.$$

This measure can be understood as follows: $\log(\frac{P(\mathbf{x})}{Q(\mathbf{x})})$ measures the log of the ratio of the probability assigned to $\mathbf{x}$ by the true distribution $P$ to the one assigned by the approximation $Q$. The entropy distance is the expected value of this log-ratio, under the true distribution, $P$. This interpretation already motivates choosing entropy distance in our application. This interpretation also shows why the entropy distance is not symmetric, i.e., $D(P\|Q)$ is not equal in general to $D(Q\|P)$. Another important property of the entropy distance function is that $D(P\|Q) \geq 0$, and equality holds if and only if $P = Q$.

Another justification of entropy distance is derived from the fact that it provides an upper bound on other distance measures. The $\mathcal{L}_1$ distance measure is defined as $\|P - Q\|_1 = \sum_{\mathbf{x}} |P(\mathbf{x}) - Q(\mathbf{x})|$.

**Theorem 3.1:** [Cover and Thomas 1991] *Let $z = \sqrt{2 \ln 2}$, and $P$ and $Q$ be two distributions on $\mathbf{U}$. Then*

$$D(P\|Q) \geq \frac{1}{z^2}\|P - Q\|_1^2.$$

Using this result, it is possible to show that the entropy distance provides a bound on all $\mathcal{L}_p$ metrics, since these are all bounded by a constant from $\mathcal{L}_1$ (note that the constant depends on the size of the space $\mathbf{U}$). We conclude that any convergence result in terms of entropy distance will immediately imply $\mathcal{L}_p$ convergence results.

Entropy distance is more sensitive than these metrics, in the following sense. Consider the situation where there are two possible outcomes, $a$ and $b$, and $P$ and $Q$ are distributions such that $P(a) = Q(a) + \epsilon$. Then $\|P - Q\|_1 = 2\epsilon$, regardless of the value of $P(a)$. On the other hand, the entropy distance from $P$ to $Q$ depends on how close $P(a)$ and $Q(a)$ are to the boundary cases, i.e., 0 or 1. To see this, note that the term $(x + \epsilon) \log \frac{x \pm \epsilon}{x}$ grows to infinity when $x$ goes to 0. Thus, the entropy distance can be arbitrarily large while $P$ and $Q$ only differ by $\epsilon$. More generally, the entropy distance is extremely sensitive to small deviations close to the boundary of the probability simplex.

The above argument indicates that convergence in terms of entropy distance is harder than convergence in the $\mathcal{L}_p$ sense. This raises the question of why we should focus on the harder notion, which can be answered in many different ways. On the axiomatic side, Shore and Johnson [1980] suggest several desired properties of approximation measures, and show that entropy distance is the only function that satisfies all of them. There are also many motivating examples. Here we briefly touch on two. The first involves compression. Suppose that we use our assessment $Q$ of a probability distribution to compress instances that are drawn in an independent manner from a distribution $P$. The length of our coded message, for a string of length $N$, is bounded below by $N(H(P) + D(P\|Q))$. Thus $D(P\|Q)$ represents the average redundancy per symbol, when compressing according to $Q$. The second example involves gambling. When gambling with allocations $Q$, on events that occur with distribution $P$, and facing (house preset) odds $O$, then wealth increases (or decreases) exponentially with exponent $N(D(P\|O) - D(P\|Q))$. (The game is repeated $N$ independent times without learning.) Thus, assuming $P$ and $O$ are fixed ($P$ by nature and $O$ by the house: their best assessment of $P$), wealth strongly depends on how close (in *entropy distance*) $Q$ is to $P$. We refer the reader to [Cover and Thomas 1991] for a discussion and a detailed analysis of these examples.

### 3.2  Treatment of the Idealized Case

In this section we make the assumption that we have an "ideal" sample sequence $\omega_N^*$ such that $\hat{P}_{\omega_N^*} = P_{B^*}$. This is clearly a strong assumption that is not even achievable for every $N$. Nonetheless, it allows us to illustrate some of the issues we encounter in the next section.

We now examine what networks are learned when the learning procedure evaluates network structures using $S_\psi(\omega_N^*, G)$. Recall that the learning procedure $Lrn_\psi(\omega_N)$ returns a network that maximizes the score. We want to ensure that we do not learn a bad network, i.e., one which is more than $\epsilon$ far from $P_{B^*}$. To do so, we have to show that for large enough $N$, all maximal networks are good. To show this, it is sufficient to show that the score of each bad network is *strictly* smaller than that of $G^*$. Thus, we are interested in showing that $S_\psi(\omega_N^*, G) < S_\psi(\omega_N^*, G^*)$ for all bad network structures $G$.

We start by examining when we would prefer one network structure to another. Straightforward algebraic manipulations of the definition of $S_\psi(\omega_N, G)$ show that:

**Lemma 3.2:** $S_\psi(\omega_N, G_1) \geq S_\psi(\omega_N, G_2)$ *if and only if*

$$D(\hat{P}_{\omega_N} \| P_{G_2,\omega_N}) - D(\hat{P}_{\omega_N} \| P_{G_1,\omega_N}) \geq \frac{\psi(N)}{N}(|G_1| - |G_2|).$$

We immediately get the following:

**Proposition 3.3:** *Let $G$ be a network such that $D(P_{B^*} \| P_{G,\omega_N^*}) > \epsilon$, and let $N$ be such that*

$$\frac{N}{\psi(N)} > \frac{|G^*| - |G|}{\epsilon}.$$

*Then $S_\psi(\omega_N^*, G) < S_\psi(\omega_N^*, G^*)$.*

Thus, for each bad network structure $G$, we can calculate a lower bound on $N$ that will ensure that $G$ is not the maximum of $S_\psi(\omega_N^*, \cdot)$. Since the size of each such bad network is at least as large as the size of the empty network, $G_\emptyset$, we easily conclude:

**Corollary 3.4:** *Let $N$ be such that*

$$\frac{N}{\psi(N)} > \frac{|G^*| - |G_\emptyset|}{\epsilon}.$$

*Then, assuming $\hat{P}_{\omega_N^*} = P_{B^*}$, we have $D(P_{B^*} \| P_{Lm_\psi(\omega_N^*)}) < \epsilon$.*

It should be noted that the bound we derive is somewhat loose: we took the naive lower bound on the size of the bad networks. Despite this looseness, and despite our nonrealistic idealizing assumption, this bound is interesting because it hints at the $\epsilon$-order of the real sample complexity.

**Proposition 3.5:** *Let $y > 4$, if $x \geq 2y \log y$, then $\frac{x}{\log x} \geq y$.*

Let $g = |G^*| - |G_\emptyset|$. We conclude that the "sample complexity in the ideal case" is in

- $O(\frac{g}{\epsilon})$, when $\psi(N) = O(1)$,
- $O(\frac{g}{\epsilon} \log \frac{g}{\epsilon})$, when $\psi(N) = O(\log N)$, and
- $O((\frac{g}{\epsilon})^{\frac{1}{1-\alpha}})$, when $\psi(N) = O(N^\alpha)$ for $\alpha > 0$.

These are consistent with one's intuition: increasing the penalty will slow the convergence to the target distribution.

### 3.3 Treatment of the Noisy Case

In the previous section we examined the sample complexity in the ideal situation where the empirical distribution we observe is exactly the underlying distribution. It is clear that in reality the sampling process introduces noise. We now show that this noise can be bounded, with high probability, for sufficiently large $N$. We can then prove an analogue of Proposition 3.3.

Our proofs rely on two results from information theory. The first is Theorem 3.1. The second is *Sanov's theorem* [Cover and Thomas 1991], which, in our setup, can be stated as follows:

**Theorem 3.6:**

$$\Pr(\{\omega_N : D(\hat{P}_{\omega_N} \| P_{B^*}) > \epsilon\}) < (N+1)^{|U|} 2^{-N\epsilon}$$

This theorem provides us with a tool for bounding the probability of "bad" samples.

With these tools in hand, we can tackle our problem. As in Proposition 3.3, we will ensure that bad networks are not learned by showing that they score strictly worse than $G^*$. We want to bound the probability of "bad" training samples $\omega_N$. Intuitively, a training sample is "bad" if there is some graph $G$ such that $P_{G,\omega_N}$ is more than $\epsilon$ far from $P_{B^*}$, yet it scores no worse than $G^*$, i.e., $S_\psi(\omega_N, G) \geq S_\psi(\omega_N, G^*)$. Using Lemma 3.2 this is equivalent to

$$D(\hat{P}_{\omega_N} \| P_{G,\omega_N}) - D(\hat{P}_{\omega_N} \| P_{G^*,\omega_N}) \leq \frac{\psi(N)}{N}(|G^*| - |G|). \tag{6}$$

It is easy to show, using Lemma 2.1, that $D(\hat{P}_{\omega_N} \| P_{B^*}) \geq D(\hat{P}_{\omega_N} \| P_{G^*,\omega_N})$. Thus, when (6) holds we also have

$$D(\hat{P}_{\omega_N} \| P_{G,\omega_N}) - D(\hat{P}_{\omega_N} \| P_{B^*}) \leq \frac{\psi(N)}{N}(|G^*| - |G|). \tag{7}$$

The key idea will be to show that if $D(\hat{P}_{\omega_N} \| P_{G,\omega_N}) - D(\hat{P}_{\omega_N} \| P_{B^*})$ is small while $D(P_{B^*} \| P_{G,\omega_N})$ is large, then it must be the case that $D(\hat{P}_{\omega_N} \| P_{B^*})$ is also large. This will allow us to apply Sanov's theorem and show that such occurrences are unlikely. The desired property could follow from a triangle inequality. However, it is well known that entropy distance does not satisfy the triangle inequality. The following lemma says that if the distributions are well-behaved, in the sense that they are not too close to the boundary of the probability simplex, then they do satisfy a similar property. For very skewed distributions, the bounds become large.

We define the following function.

$$e(a, b, c, m) = \frac{1}{2} z b^{\frac{1}{2}} \frac{\frac{z}{m}\left(a + zb^{\frac{1}{2}}c\right)^{\frac{1}{2}}}{1 - \frac{z}{m}\left(a + zb^{\frac{1}{2}}c\right)^{\frac{1}{2}}} + a$$

**Lemma 3.7:** *Let $P, Q, R$ be probability distributions over $U$ such that $D(P\|Q) - D(P\|R) \leq a$, $D(P\|R) \leq b$, and $\max_x \left|\log \frac{R(x)}{Q(x)}\right| \leq c$. Let $m(R) = \min_x R(x)$. Also assume that $a, b, c$ above satisfy $z\left(a + zb^{\frac{1}{2}}c\right)^{\frac{1}{2}} < m(R)$. Then $D(R\|Q) < e(a, b, c, m(R))$.*

We now face a technical problem. The bound in this lemma is based on assuming $\max_{\mathbf{x}} \left|\log \frac{R(\mathbf{x})}{Q(\mathbf{x})}\right| \leq c$. The applications of this bound will set $R$ to $P_{B^*}$, which we assume to be a constant, but it will also set $Q$ to $P_{G,\omega_N}$, over which we do not have much control.

Note that the probability $P_{G,\omega_N}$ depends on the sample $\omega_N$. We now characterize these samples that generate distributions that do not satisfy the requirements of Lemma 3.7. Let

$$Skew(c) = \{\omega_N : \exists G, \max_{\mathbf{x}} \left|\log \frac{P_{B^*}(\mathbf{x})}{P_{G,\omega_N}(\mathbf{x})}\right| > c\}$$

be the set of "skewed" samples. We now show that, for an appropriately chosen $c$, this set has small probability when $N$ is sufficiently large. $c$, here, depends on how close the target distribution is to the boundary of the probability simplex. From here on, let $m = \min_{\mathbf{x}} P_{B^*}(\mathbf{x})$.

280    Friedman and Yakhini**Lemma 3.8:**

$$\Pr(Skew(2n\log\frac{1}{m})) < (N+1)^{\|\mathbf{U}\|}2^{-N\left(\frac{(1-m)m}{2}\right)^2}.$$

We are now ready to apply Sanov's theorem and get a convergence rate.

**Theorem 3.9:** *Fix $a$, $b$, and $N$ such that*

$$\frac{N}{\psi(N)} \geq \frac{|G^*| - |G_\emptyset|}{a},$$

*and $z(a + zb^{\frac{1}{2}}2n\log\frac{1}{m})^{\frac{1}{2}} < m$. Let $\epsilon = \epsilon(a, b, 2n\log\frac{1}{m}, m)$ and let $\delta = (N+1)^{\|\mathbf{U}\|}(2^{-Nb} + 2^{-N\left(\frac{(1-m)m}{2}\right)^2})$. Then*

$$\Pr(D(P_{B^*}\|P_{Lrn_\psi(\omega_N)}) > \epsilon) < \delta.$$

**Proof:** There are two generic classes of samples for which $Lrn_\psi$ might return a bad network. The first is when $\omega_N$ is skewed, i.e., in $Skew(2n\log\frac{1}{m})$. The second is when $\omega_N$ is not skewed, but there is a network structure $G$ such that $P_{G,\omega_N}$ is far from $B^*$, yet $G$ scores no worse than $G^*$'s. Lemma 3.8 takes care of the first. We now examine the second case. Let $\mathcal{G} = \mathcal{G}(\omega_N, \epsilon) = \{G : D(P_{B^*}\|P_{G,\omega_N}) > \epsilon\}$ be the set of "bad" network structures. Let

$$E = \{\omega_N : \exists G \in \mathcal{G}, S(\omega_N, G) \geq S(\omega_N, G^*)\}.$$

We now want to bound the probability of $E - Skew(2n\log\frac{1}{m})$. Using (7) we observe that $E \subset E'$ where $E'$ is the set

$$\left\{\omega_N : \exists G \in \mathcal{G}_\epsilon, D(\hat{P}_{\omega_N}\|P_{G,\omega_N}) - D(\hat{P}_{\omega_N}\|P_{B^*}) \leq a\right\}.$$

We also define

$$F = \{\omega_N : D(\hat{P}_{\omega_N}\|P_{B^*}) > b\}.$$

Clearly,

$$E' - Skew(2n\log\frac{1}{m}) \subseteq F \cup (E' - (F \cup Skew(2n\log\frac{1}{m}))).$$

We easily bound $F$ using Theorem 3.6 (Sanov). Applying Lemma 3.7 with $P = \hat{P}_{\omega_N}$, $R = P_{B^*}$, and $Q = P_{G,\omega_N}$, we conclude that the set $E' - (F \cup Skew(2n\log\frac{1}{m}))$ is empty. ∎

This theorem captures the interaction between the approximation error, the confidence bound, and the sample size. However, it relates these terms in an unwieldy form. Given $\epsilon$ and $\delta$, we have to choose values of $a$ and $b$ so as to obtain a reasonable sufficient value of $N$.

Finding these values of $a$ and $b$ we can get the following asymptotic results. Recall that we define $g = |G^*| - |G_\emptyset|$.

**Theorem 3.10:** *Let $F(x)$ be the inverse of $\frac{x}{\log(x)}$, $x > 2$. Let $N_\psi(\epsilon, \delta)$ be the sample complexity of $Lrn_\psi()$. If $\psi(N) \in O(\log(N))$, then*

$$N_\psi(\epsilon, \delta) \in O(F(((\|\mathbf{U}\| + \log\frac{1}{\delta})(\frac{1}{\epsilon})^{\frac{4}{3}}(\frac{1}{m})^2)),$$

*If $\psi(N) = N^\alpha$ for $0 < \alpha \leq \frac{1}{3}$, then*

$$N_\psi(\epsilon, \delta) \in O(F(((\|\mathbf{U}\| + \log\frac{1}{\delta})^{\frac{2}{3-\alpha}}(\frac{1}{\epsilon})^{\frac{4}{3-\alpha}}g^{\frac{1}{1-\alpha}}(\frac{1}{m})^{\frac{2}{1-\alpha}})).$$

*Both asymptotic statements are as $(\epsilon, \delta, m) \to 0$ such that $\epsilon^2 \leq m/(n^{\frac{1}{2}}\log^{\frac{1}{2}}\frac{1}{m})$.*

We note that $F(\cdot)$ does not have a closed-form. However, Proposition 3.5 implies that $O(F(x)) \subseteq O(x\log x)$ as $x \to \infty$. Moreover, we note that these results describe the asymptotic order of the sample complexity, with respect to all the parameters involved. Often we can assume that $\mathbf{U}$, $n$, $g$, and $m$ are fixed, and examine the behavior of $N$ with respect to $\epsilon$ and $\delta$. In particular, we get the upper bound described in the introduction (for logarithmic $\psi(N)$):

**Corollary 3.11:** *Fix a universe $\mathbf{U}$ and minimal skewness $m$. Then there exists a function $N(\epsilon, \delta)$ satisfying*

$$N(\epsilon, \delta) \in O\left((\frac{1}{\epsilon})^{\frac{4}{3}}\log\frac{1}{\epsilon}\log\frac{1}{\delta}\log\log\frac{1}{\delta}\right)$$

*such that if $\min_\mathbf{x} P_{B^*}(x) > m$, then whenever $N > N(\epsilon, \delta)$,*

$$\Pr(D(P_{B^*}\|P_{Lrn_{\frac{1}{2}\log N}(\omega_N)}) > \epsilon) < \delta.$$

It is interesting to note that our upper bound on the sample complexity does not depend on $|G^*|$ when $\psi(N) \leq \log N$. This suggests that the main factor that determines the convergence rate for these penalty factors is the sample noise. On the other hand, our result also suggests the structure size does play a role when the penalty is polynomial in $N$. These statements, however, cannot be made precise using our upper-bound results which are not necessarily tight.

### 3.4 Behavior in the Limit

The sample complexity results show how fast the learning procedure converges to the target distribution. So far we did not examine the network structure learned by the procedure. The use of Bayesian networks for various tasks exploits the independences encoded by the network structure. As a general rule, a structure that makes stronger independence assumptions will lead to more efficient computations. Thus, ideally, we would prefer to learn a minimal network that describes the target distribution. This suggests that we should distinguish two criteria for learning procedures. When a learning procedure $Lrn_\psi$ converges almost surely to the target distribution, we say that it is *(asymptotically) consistent*. A stronger requirement is that the network structure recovered by $Lrn_\psi$ is also a minimal network for sufficiently large $N$. We call such procedures *(asymptotically) minimal*.

We now define these notions formally. Let $\omega$ be an infinite sequence of instances of $\mathbf{U}$. We define $\omega_N$ to be the sequence of the first $N$ elements of $\omega$. We say that a learning procedure $Lrn_\psi$ is *asymptotically consistent* if

$$\Pr(\{\omega : \lim_{N\to\infty} D(P_{B^*}\|P_{Lrn_\psi(\omega_N)}) = 0\}) = 1.$$

Using our convergence rate results, we can easily show that all reasonable penalty functions $\psi(N)$ lead to consistent learning procedures:

**Corollary 3.12:** *If $\lim_{N\to\infty}\frac{\psi(N)}{N} = 0$, then $Lrn_\psi$ is asymptotically consistent.*

**Proof:** Use Theorem 3.9 and Borel-Cantelli's lemma ([Feller 1957]). ∎

This consistency result for MDL learning is by no means new. Barron and Cover [1991] treat such questions in a general setting, but without obtaining a confidence rate. When $\psi(N) = \frac{1}{2}\log N$, consistency follows from Schwarz's result [1978] and the well-known consistency properties of Bayesian learning.

We now turn to asymptotical minimality. Recall that we assume that $G^*$ is the minimal network structure that can describe the target distribution. More precisely, let

$$\mathcal{G}^* = \{G : \text{there is a network}, B = \langle G, \Theta \rangle \text{ s.t. } P_B = P_{B^*}\}.$$

We assume that $|G^*| = \min_{G \in \mathcal{G}^*} |G|$. We say that a consistent learning procedure is *asymptotically minimal* if it eventually returns structures that are not more complicated than $G^*$, i.e.,

$$\Pr(\{\omega : |Gph(Lrn_\psi(\omega_N))| > |G^*| \text{ infinitely often }\}) = 0.$$

Thus, a minimal learning procedure will, almost surely, eventually find a minimal network structure for the target distribution. To see this notice that an asymptotically consistent procedure will always select graphs that are at least as large as $G^*$ from some point onward: Suppose that $|G| < |G^*|$. From the minimality of $G^*$ it follows, using a compactness argument, that $\inf_\Theta D(P_{B^*} \| P_{\langle G,\Theta \rangle}) > 0$. Thus, consistency implies that, almost surely, for sufficiently large $N$, $G$ would not be learned by $Lrn_\psi$.

**Theorem 3.13:** *Let $\psi(N)$ be such that $\lim_{N \to \infty} \frac{\log N}{\psi(N)} = \lim_{N \to \infty} \frac{\psi(N)}{N} = 0$. Then $Lrn_\psi$ is asymptotically minimal.*

**Proof:** Fix a graph, $G$, with $|G| > |G^*|$, then

$$\begin{aligned} A_N &= \{\omega : Gph(Lrn_\psi(\omega_N) = G\} \\ &\subseteq \{\omega : D(\hat{P}_{\omega_N} \| P_{B^*}) \geq \frac{\psi(N)}{N}(|G| - |G^*|)\}. \end{aligned}$$

Where the inclusion holds by Lemmas 3.2 and 2.1. Apply Sanov to obtain

$$\Pr(A_N) \leq (N+1)^{\|\mathbf{U}\|} 2^{-\psi(N)(|G|-|G^*|)}.$$

Therefore $\Pr(\{\omega : \omega \in A_N \text{ infinitely often }\}) = 0$ (Borel-Cantelli and our assumption on $\psi(N)$). Since there are only finitely many graph structures, the proof is complete. ∎

This raises the question of whether $Lrn_\psi$ is asymptotically minimal when $\psi(N) = \frac{1}{2}\log N$. We suspect that it is indeed minimal in this case, but cannot formally prove it. (It is clear that we cannot use the technique we just discussed to prove such a result.) We also suspect that when $\psi(N) = c$, the learning procedure is not minimal. We leave these issues for future work.

## 4 Sub-Sampling Strategies in Learning

We now discuss the applications of sample complexity results to speeding up learning processes. As described in Section 2, we evaluate candidate networks by computing the description length. This involves evaluating the complexity of the candidate and $LL_{\omega_N}(G)$. The latter term can be decomposed according to the structure of the network, as shown in (5). We therefore need to evaluate expressions of the form $H_{\hat{P}_{\omega_N}}(X_i|\Pi_{X_i}) = H_{\hat{P}_{\omega_N}}(X_i, \Pi_{X_i}) - H_{\hat{P}_{\omega_N}}(\Pi_{X_i})$.

In most current learning systems, these expressions are evaluated by scanning through the training data and evaluating the joint frequency of $X$ and $\Pi_X$. This, however, is extremely inefficient when we have a large number of instances in our training data. We now examine an approach that uses *sub-sampling* to evaluate such expressions. The rough idea is as follows. When evaluating $H(X|\Pi_X)$, we will randomly sample $k$ instances from the data, and evaluate the conditional entropy on these instances. If $k$ is much smaller than $N$, then this considerably reduces the evaluation time. However, we must ensure that by ignoring most of the training data we have not introduced a large error.

The next result bounds the error in estimating $H(X)$ from sample data. $H_{\hat{P}_{\omega_N}}(X)$ denotes the empirical estimate of the entropy of $X$, and $H(X)$ is the entropy of $X$ under the underlying distribution.

**Proposition 4.1:** *Let $X$ a random variable taking finitely many values, with probabilities satisfying $\min_{v \in val(X)} P(X = v) \geq m$. Let $\Pr$ be the corresponding product measure. Let $\epsilon < 1/4$. Then*

$$\Pr(|H_{\hat{P}_{\omega_N}}(X) - H(X)| > \epsilon) \leq (N+1)^{\|X\|} 2^{-N\epsilon^2 \frac{1}{3\log\frac{1}{m}}}.$$

This suggests a modified learning procedure that uses sub-sampling. This procedure is called with parameters $\epsilon, \delta$. Each candidate's log-likelihood score is now evaluated as follows. The procedure computes the sample size $N_{X_i}$ necessary to compute the required entropy term (as mentioned above) with bound $(\frac{\epsilon}{2n}, \frac{\delta}{2n})$, according to Proposition 4.1. Then, it takes a sub-sample of size $N_{X_i}$ from $\omega_N$ and computes $H(X_i|\Pi_{X_i})$ using the observed frequencies. This clearly leads to saving in practice when the training data size is very large.

We note that ideas, based on similar intuition, were pursued in the context of learning decision trees in [Musick, Catlett, and Russell 1992]. However, the methods used in learning decision trees are quite different than the ones necessary for Bayesian networks.

## 5 Discussion

In this paper, we derive an upper bound on the sample complexity of learning BNs using the MDL score. As stated in Corollary 3.11, that sample complexity satisfies

$$N(\epsilon, \delta) \in O\left((\frac{1}{\epsilon})^{\frac{4}{3}} \log \frac{1}{\epsilon} \log \frac{1}{\delta} \log \log \frac{1}{\delta}\right)$$

We note that the bound we derive is quite loose. In particular, the lemma that largely determines the convergence rate (Lemma 3.7) does not exploit the structure of the network, but only examines the properties of entropy distance between measures. We hope that a more refined analysis, taking this structure into account, will yield tighter bounds on the sample complexity. The discussion in Section 3.2 suggests that we cannot hope for an $\epsilon$-asymptotic behavior which is better than $O(\frac{1}{\epsilon}\log(\frac{1}{\epsilon}))$. This rate is an experimentally observed one. Figure 1 shows the learning curves for three different networks with the cross-entropy error scaled by $\frac{N}{\log N}$. (These results are taken from [Friedman and Goldszmidt 1996], where details on the experimental

282  Friedman and Yakhini

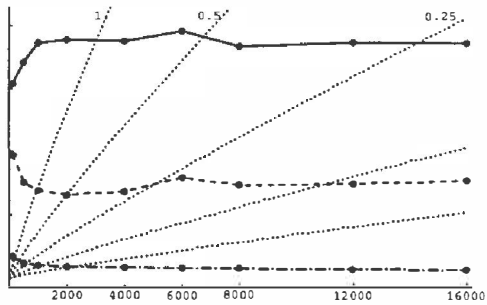

Figure 1: Learning curves for learning 3 networks. The horizontal axis measures the number of samples, $N$. The vertical axis measures the average cross-entropy error from 10 experiments multiplied by $\frac{N}{\log N}$. The dotted diagonal lines are lines of constant error. These results are taken from [Friedman and Goldszmidt 1996]. The learning curves are of Alarm (solid line), CTS (dashed line), and TJ (dot-dash line).

setup can be found.) As we can see, the learning curves are roughly constant from some point onward, suggesting that the error is asymptotically proportional to $\frac{\log N}{N}$.

The only other work we are aware of, that examines the issue of sample complexity of learning BNs, is that Höffgen [1993]. He examines the sample complexity of learning networks of Boolean variables, where each variable has at most $k$ parents. He provides an upper bound in $O((\frac{1}{\epsilon}\log\frac{1}{\epsilon})^2 \log\frac{1}{\delta})$. This bound is much worse than ours in terms of $\epsilon$. One of the constants involved in his upper bound is $2^k$. Thus, it is not surprising that when we allow arbitrary networks (i.e., $k = n$), we get the constant $2^n = ||\mathbf{U}||$ in our analysis. Höffgen also examines the computational aspects of learning BNs and shows that for $k > 1$, the task of searching for the best scoring structure is NP-hard.

Höffgen's analysis involves tilting the empirical distributions, towards the uniform distribution, with a low weight. This technique allows him to derive a bound which is independent of $m$, the skewness of the target distribution. We are currently considering using a similar technique to reduce the $m$-order of our results.

Another related issue is Bayesian approaches to learning BNs [Cooper and Herskovits 1992; Heckerman, Geiger, and Chickering 1995; Heckerman 1995]. In these approaches we assume that there is a prior over possible network structures and their associated parameters. Learning is done by selecting a structure with maximum *posterior* probability, given the data. This form of learning generalizes the MDL learning we examined here, since, roughly speaking, the MDL metric essentially chooses a particular prior. It is known [Schwarz 1978] (see [Heckerman 1995]) that if the prior is not extreme, under a suitable definition, then the posterior $P(G|\omega_N)$ is in $O(2^{-S_\psi(\omega_N,G)})$ as $N \to \infty$. Thus, in the limit, candidates' score similarly under the two metrics. This suggests that sample complexity questions for Bayesian learning method may be related to the ones we discussed here. However, we do not know the rate of convergence of the Bayesian posterior to the MDL score. In particular, a highly biased prior can slow the convergence rate of the Bayesian learner. Nonetheless, we believe that the techniques we introduced provide a good initial point in the examination of the sample complexity of Bayesian learning.

We also discuss questions of asymptotic minimality and of sample complexity for polynomial penalty functions. The answers we give here are far from complete, and we hope to investigate some of the gaps in the future.

Acknowledgments

The authors are grateful to Moises Goldszmidt, Daphne Koller, an anonymous reviewer, and especially Joe Halpern for comments on a previous draft of this paper and useful discussions relating to this work. Parts of this work were done while the first author was at Rockwell Science Center. The first author was also supported in part by an IBM Graduate fellowship and NSF Grant IRI-95-03109.

## References

Akaike, H. (1974). A new look at the statistical identification model. *IEEE Trans. on Auto. Control 19*, 716–723.

Barron, A. R. and T. M. Cover (1991). Minimum complexity density estimation. *IEEE Trans. on Info. Theory 37*, 1034–1054.

Cooper, G. F. and E. Herskovits (1992). A Bayesian method for the induction of probabilistic networks from data. *Machine Learning 9*, 309–347.

Cover, T. M. and J. A. Thomas (1991). *Elements of Information Theory*. Wiley.

Feller, W. (1957). *An Introduction to Probability Theory and its Applications* (2nd ed.), Volume 1. Wiley.

Friedman, N. and M. Goldszmidt (1996). Learning Bayesian networks with local structure. In *UAI '96*.

Friedman, N. and Z. Yakhini (1996). On the sample complexity of learning Bayesian networks. Available via WWW at http://robotics.stanford.edu/users/nir.

Heckerman, D. (1995). A tutorial on learning Bayesian networks. Technical Report MSR-TR-95-06, Microsoft Research.

Heckerman, D., D. Geiger, and D. M. Chickering (1995). Learning Bayesian networks: The combination of knowledge and statistical data. *Machine Learning 20*, 197–243.

Höffgen, K. L. (1993). Learning and robust learning of product distributions. In *COLT '93*, pp. 77–83.

Kullback, S. and R. A. Leibler (1951). On information and sufficiency. *Ann. Math. Stat. 22*, 76–86.

Lam, W. and F. Bacchus (1994). Learning Bayesian belief networks. An approach based on the MDL principle. *Computational Intelligence 10*, 269–293.

Musick, R., J. Catlett, and S. Russell (1992). Decision theoretic subsampling for induction on large databases. In *ML '93*.

Pearl, J. (1988). *Probabilistic Reasoning in Intelligent Systems*. Morgan Kaufmann.

Rissanen, J. (1989). *Stochastic Complexity in Statistical Inquiry*. World Scientific Publ.

Schwarz, G. (1978). Estimating the dimension of a model. *Ann. of Stat. 6*, 461–464.

Shore, J. E. and R. W. Johnson (1980). Axiomatic derivation of the principle of maximum entropy and the principle of minimimum cross-entropy. *IEEE Trans. on Info. Theory 26*, 26–37.

Suzuki, J. (1993). A construction of Bayesian networks from databases based on an MDL scheme. In *UAI '93*, pp. 266–273.